%% file: main.tex
\pdfoutput=1

\documentclass[11pt]{article}

\usepackage{tcolorbox}  
\tcbuselibrary{breakable}
\usepackage{booktabs}
\usepackage{makecell} 
\tcbset{fontupper=\small}
\usepackage[final]{acl}
\usepackage{enumitem}
\usepackage{times}
\usepackage{latexsym}
\usepackage{xspace}
\usepackage{tabularx}
\usepackage[T1]{fontenc}

\usepackage[utf8]{inputenc}

\usepackage{microtype}

\usepackage{inconsolata}
\usepackage{titlesec}

\usepackage{graphicx}
\usepackage{amssymb}
\usepackage{multirow}
\usepackage{algorithm}
\usepackage{algorithmic}
\usepackage{amsmath}
\usepackage{booktabs}
\usepackage{xcolor}
\usepackage{subfigure}
\newcommand{\name}{MultiDx\xspace}
\title{MultiDx: A Multi-Source Knowledge Integration Framework \\towards Diagnostic Reasoning}

\author{
    Yimin Deng\textsuperscript{1,2}\thanks{Work was conducted at Tencent Jarvis Lab.},
    Zhenxi Lin\textsuperscript{3},
    Yejing Wang\textsuperscript{2},
    Guoshuai Zhao\textsuperscript{1}$^{\dagger}$,
    Pengyue Jia\textsuperscript{2},
      \textbf{Zichuan Fu}\textsuperscript{2},
    \\
    \textbf{Derong Xu}\textsuperscript{2},
     \textbf{Yefeng Zheng}\textsuperscript{4},
     \textbf{Xiangyu Zhao\textsuperscript{2}$^{\dagger}$},
     \textbf{Li Zhu\textsuperscript{1}$^{\dagger}$}, 
     \textbf{Xian Wu\textsuperscript{3}$^{\dagger}$},
     \textbf{Xueming Qian}\textsuperscript{1}
    \\
    \textsuperscript{1}Xi'an Jiaotong University, \textsuperscript{2}City University of Hong Kong,\\
    \textsuperscript{3}Tencent Jarvis Lab,
    \textsuperscript{4} Westlake University
    \\
  \small{
  \texttt{
  \href{mailto:dymanne@stu.xjtu.edu.cn}{dymanne@stu.xjtu.edu.cn},
    \href{mailto:guoshuai.zhao@xjtu.edu.cn}{guoshuai.zhao@xjtu.edu.cn},
     \href{mailto:xianzhao@cityu.edu.hk}{xianzhao@cityu.edu.hk}}}
     \\
     \small{
      \texttt{
    \href{mailto:zhuli@xjtu.edu.cn}{zhuli@xjtu.edu.cn},
    \href{mailto:kevinxwu@tencent.com}{kevinxwu@tencent.com}
  }}
  } 

\begin{document}
\maketitle
\begingroup
\renewcommand\thefootnote{\relax}
\footnotetext{$^{\dagger}$  Corresponding authors.}
\endgroup
\begin{abstract}
  Diagnostic prediction and clinical reasoning are critical tasks in healthcare applications. While Large Language Models~(LLMs) have shown strong capabilities in commonsense reasoning, they still struggle with diagnostic reasoning due to limited domain knowledge.~Existing approaches often rely on internal model knowledge or static knowledge bases, resulting in knowledge insufficiency and limited adaptability, which hinder their capacity to perform diagnostic reasoning.~Moreover, these methods focus solely on the accuracy of final predictions, overlooking alignment with standard clinical reasoning trajectories.~To this end, we propose \name{}, a two-stage diagnostic reasoning framework that performs differential diagnosis by analyzing evidence collected from multiple knowledge sources.~Specifically, it first generates suspected  diagnoses and reasoning paths by leveraging knowledge from web search, SOAP-formatted case, and clinical case database.~Then it integrates multi-perspective evidence through matching, voting, and differential diagnosis to generate the final prediction.~Extensive experiments on two public benchmarks demonstrate the effectiveness of our approach. The code is available at \url{https://github.com/Applied-Machine-Learning-Lab/ACL2026-MultiDx}.

\end{abstract}
\input{1Introduction}

\input{3Methods}
\input{4Experiments}

\input{2RelatedWork}

\input{5Conclusion}

\input{6Limitations}

\bibliography{main}
\clearpage
\appendix

\input{7Appendix}

\end{document}

%% file: 1Introduction.tex
\section{Introduction}

\begin{figure}[t]
    \centering
    \centering
   \includegraphics[height=0.4\textwidth]{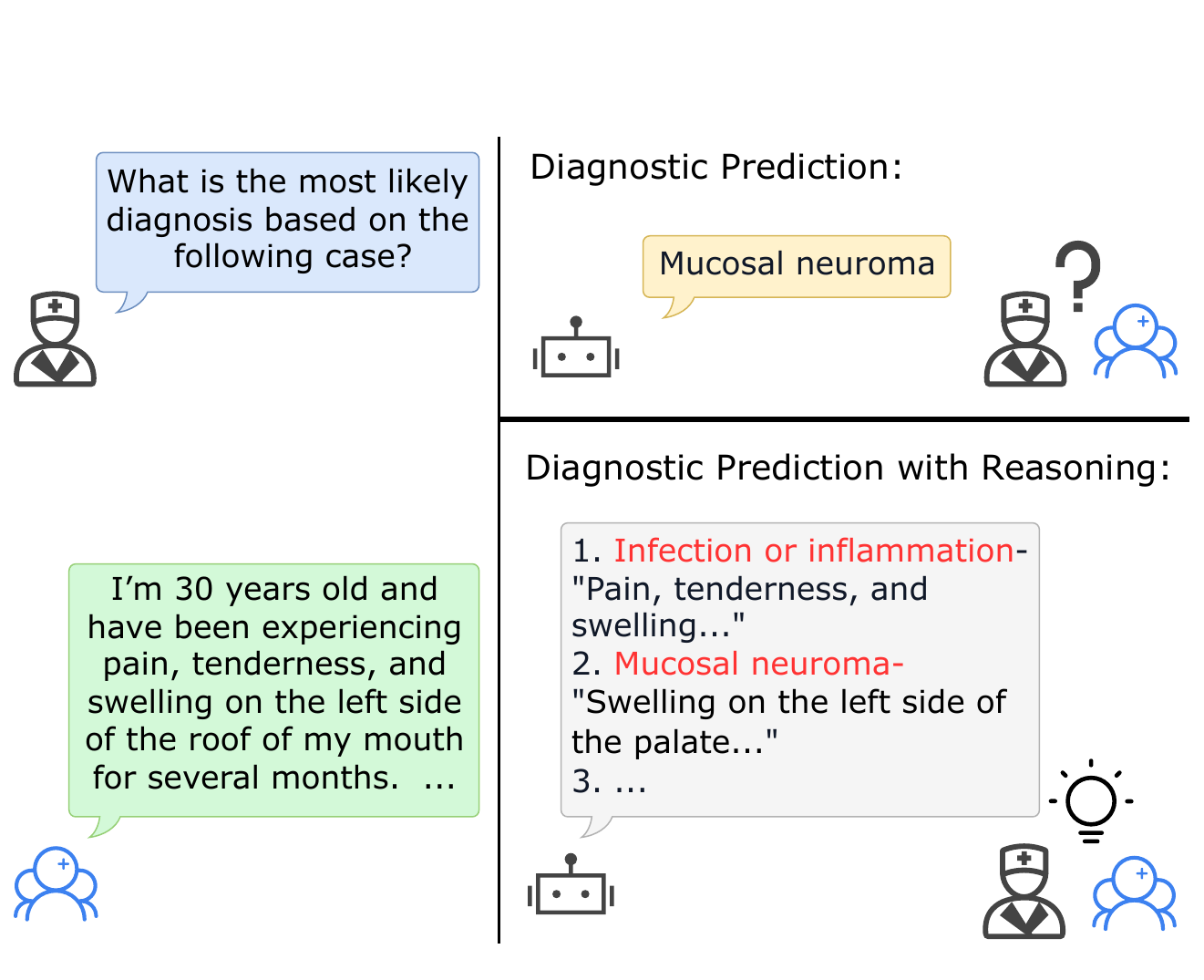}
    \caption{An example of diagnosis reasoning.
    }
    \label{setting}
\end{figure}

Diagnostic reasoning is a critical downstream task in clinical applications~\cite{patel2005thinking,lucas2024reasoning}. It aims to integrate information from clinical case reports (e.g., patient symptoms, test results) to establish a diagnosis (or identify the disease). Beyond achieving the correct diagnostic prediction, it is equally important to ensure that the reasoning process adheres to established medical standards~\cite{wu2025medcasereasoning}.
As illustrated in Figure~\ref{setting}, diagnostic predictions without reasoning can be difficult to verify or justify, potentially undermining trust. In contrast, predictions with structured reasoning not only facilitates patient understanding of their condition but also provides clinicians with verifiable and trustworthy decision support. Achieving interpretable diagnostic predictions forms the foundation of reliable and responsible  AI-assisted healthcare systems.

In recent years, Large Language Models (LLMs) have been widely applied to medical reasoning~\cite{tang2025medagentsbench,chen2025medbrowsecomp}. Compared to commonsense reasoning, medical reasoning heavily relies on accurate and comprehensive domain-specific knowledge.~Existing approaches incorporate medical knowledge in two main ways.~One the one hand, some methods rely solely on the internal knowledge of LLMs.~For instance, MedAgents~\cite{tang2024medagents} introduces a training-free collaborative framework in which LLMs act as medical specialists, and MDAgents~\cite{kim2024mdagents} proposes an adaptive strategy that dynamically assigns agents based on task complexity.
On the other hand, recent methods attempt to enhance reasoning capabilities by leveraging static knowledge bases. Specifically, Medaide~\cite{wei2024medaide} integrates information from multiple domain-specific databases to support medical reasoning.
ConfAgents~\cite{zhao2025confagents} adopts an adaptive and cost-efficient framework that retrieves relevant information from an external document repository.
MMedAgent-RL~\cite{xia2025mmedagent} leverages reinforcement learning to learn knowledge from databases.
However, current approaches often lack effective mechanisms to integrate diverse and dynamic external knowledge sources, which limits their ability to provide sufficient support for clinical reasoning, especially in unseen or rare cases.

Despite recent progress, generating high-quality reasoning trajectories remains a significant challenge.~Most existing approaches are evaluated on benchmark datasets such as MedQA and PubMedQA~\cite{jin2021disease,jin2019pubmedqa}, which primarily focus on the correctness of the final answer, rather than the quality of the diagnostic reasoning process.
As a result, the diagnosis reasoning process are often under-specified, lacking a coherent chain of medical evidence. This not only reduces the explainability of the system but also undermines its verifiability from a clinical perspective. Without a well-grounded reasoning path, it becomes difficult for clinicians to trust the model’s recommendations or to cross-check them with medical knowledge. Consequently, these limitations restrict the applicability of current models in real-world clinical settings, where rigorous justification and interpretability are essential.

Therefore, it is crucial to incorporate comprehensive medical knowledge and to construct high-quality reasoning trajectories.~To address these challenges, we propose \name{}, a two-stage diagnostic reasoning framework enhanced with multi-source knowledge. 
In the first stage, the model leverages multiple knowledge sources, including web search results, structured case reports, and medical case database, to generate disease lists along with corresponding reasoning paths.
In the second stage, the model performs disease matching, voting, and differential diagnosis to produce the final prediction and its associated reasoning trace.
To ensure consistency with clinical practice, our method aligns with two core steps of the standard diagnostic workflow: generating a suspected disease list, followed by differential diagnosis through multi-perspective reasoning. Moreover, we incorporate the SOAP~\cite{podder2021soap}, a widely accepted format in clinical practice, to help the model effectively organize clinical case.

To summarize, our contributions are as follows:
\begin{itemize}[leftmargin=*,nolistsep]

\item We propose \name{}, a two-stage diagnostic reasoning framework that integrates multi-source knowledge while explicitly aligning with standard clinical practices.
\item By incorporating diverse knowledge sources, our model can analyze clinical cases from multiple perspectives and generate comprehensive, evidence-based reasoning trajectories.
\item We conduct thorough experiments on two public diagnostic reasoning benchmarks, demonstrating that our method yields both high-quality reasoning paths and accurate predictions.

\end{itemize}

%% file: 3Methods.tex
\section{Methods}
    \begin{figure*}[t]
    \centering
    \resizebox{1\linewidth}{!}{
    \includegraphics[height=0.7\textwidth]{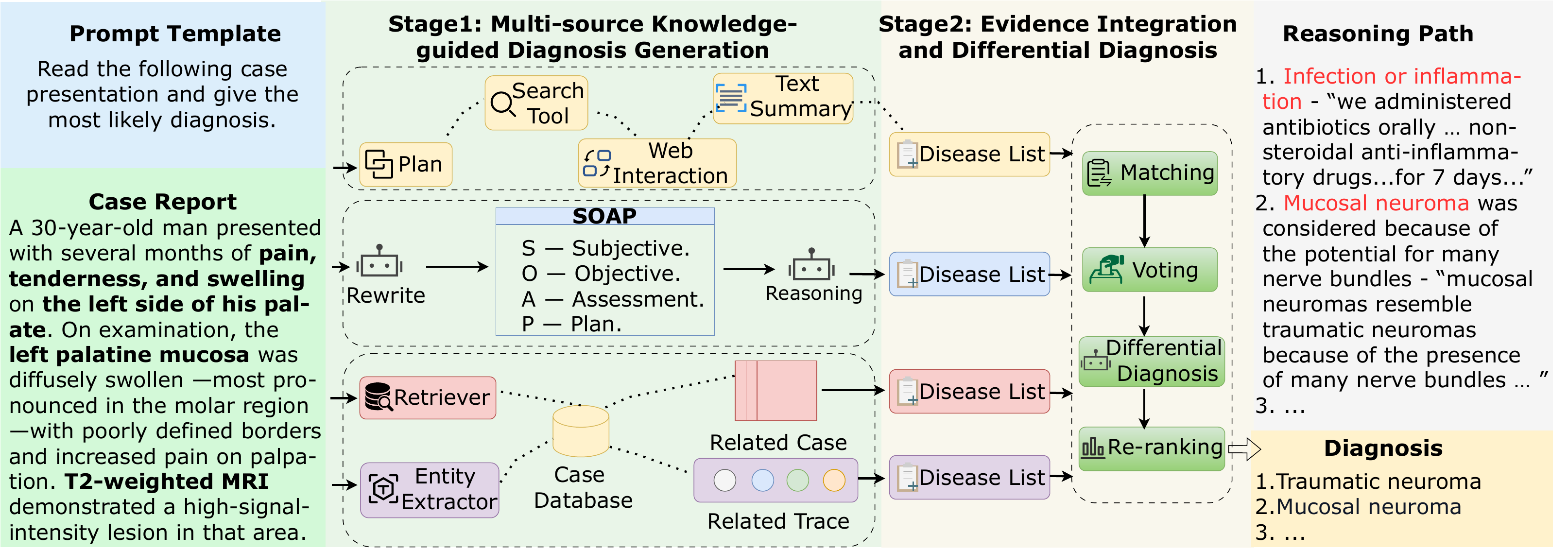}}
    \caption{The overall architecture of \name{}. 
    }
    \label{model}
\end{figure*}
In this section, we provide a comprehensive overview of our proposed diagnostic reasoning framework \name{}. We first introduce the problem definition of the diagnostic reasoning task in section~\ref{formulation}, and then present the overall architecture of \name{} in section~\ref{overall}, followed by a detailed description of each component in the two-stage pipeline in section~\ref{stage1} and section~\ref{stage2}.
\subsection{Problem Formulation}\label{formulation}
The diagnostic reasoning task aims to generate a diagnostic prediction along with clinically sound reasoning paths based on a patient’s case report. Formally, each data instance is represented as a tuple $\{C, R, D\}$, where $C$ denotes the case report, $R$ the reasoning path, and $D$ the final diagnosis. As shown in Figure \ref{model}, given a case report $C$ that contains relevant clinical information~(e.g., subjective symptoms and objective examination results), the model is expected to produce a reasoning path $R$ that shows the step-by-step diagnostic process based on medical evidence, and a diagnosis $D$. 

\subsection{Overall Framework}\label{overall}
In this section, we introduce the overall architecture of our model. As shown in Figure \ref{model}, our model consists of two stages, starting with \textit{Multi-source Knowledge-guided Diagnosis Generation}, and followed by \textit{Evidence Integration and Differential Diagnosis}. Specifically, in the first stage, the model retrieves knowledge from multiple sources: web search, structured case, case database, and generates a list of suspected diseases for each source. In the second stage, the model performs disease matching, voting, and differential diagnosis to integrate the results into a final diagnosis.
\subsection{Multi-source Knowledge-guided Diagnosis Generation}\label{stage1}
Unlike commonsense or math reasoning, medical reasoning relies on complete domain-specific knowledge to analyze the associations between symptoms or examination findings and potential diseases. In this section, we present how evidence is gathered from multiple complementary knowledge sources, each contributing to the generation of a list of suspected diseases.

\subsubsection{SOAP}
The natural language case reports received by the model are often unstructured, mixing subjective complaints and objective findings in a free-form narrative.~It limits the model’s ability to process different types of clinical information effectively~\cite{pearce2016essential,seo2016pilot}.
To address these limitations and facilitate diagnostic reasoning, we adopt the SOAP format,  which is widely used in clinical practice~\cite{podder2021soap}.~It divides the original clinical case into four components:

- \textbf{Subjective}: Patient-reported symptoms.

- \textbf{Objective}: Observed signs, test results, and physical examination findings.

- \textbf{Assessment}: Generation of a list of candidate diseases based on subjective and objective data.

- \textbf{Plan}: Suggestion of potential diagnostic or treatment steps tailored to the patient's condition.

Formally, given an instruction prompt $I_\text{toSOAP}$ for text structuring and an unstructured clinical case $C$, we utilize a LLM to generate its structured representation $C_\text{SOAP}$ in SOAP format:
\begin{equation}
C_\text{SOAP} = \text{LLM}(I_\text{toSOAP},C)
\label{eq:soap_encoding}
\end{equation}

Since the input to our system is the original case prompt, which contains only patient-reported symptoms and objective findings without any preliminary diagnosis or treatment plan, we use the SOAP format solely to structure this potentially noisy input into clinically standardized components. In our generated SOAP record, both Assessment and Plan are explicitly marked as absent.

This structured representation is then used as the input for diagnostic reasoning agent. Given a clinical inference prompt $I_\text{SOAP}$, the LLM generates a disease list $H_\text{SOAP}$:
\begin{equation}
H_\text{SOAP} = \text{LLM}(I_\text{SOAP},C_\text{SOAP})
\label{eq:soap_inference}
\end{equation}

By structuring inputs according to SOAP, we align the reasoning process with clinical standard. 

\subsubsection{Medical Case Database}
Diagnostic reasoning follows clinical standards in real-world scenarios. It has been shown that adherence to these guidelines improves diagnostic performance. To enable the model to learn diagnostic reasoning patterns that follow clinical workflows, we incorporate hierarchical retrieval-augmented generation using a database of clinician-annotated cases. 

Formally, the database \( \mathcal{G} = \{(C_i, R_i, D_i)\}_{i=1}^N \), where each instance consists of a clinical case description \( C_i \), the associated reasoning trace \( R_i \), and the final diagnostic prediction \( D_i \).
Then we perform hierarchical retrieval, including retrieve similar cases based on textual similarity, and retrieve reasoning steps based on extracted medical entity.

\textbf{Related Case.}
Given a input case \( C \), we use a BM25 retriever to identify the top-\( k \) most similar cases from the corpus based on textual similarity:
\begin{equation}
\text{TopK}(C) = \text{arg top-}k_{(C_i, R_i, D_i) \in \mathcal{D}} \ \text{BM25}(C, C_i)
\end{equation}

The retrieved tuples are then concatenated into the prompt as additional context. Specifically, we build an augmented input by including the retrieved exemplars followed by the current case \( C \):
\begin{equation}
H_{case} = \text{LLM}(I_{case}, \text{TopK}(C), C)
\end{equation}

Here, \( I_{case} \) denotes a reasoning instruction, and \( H_{case} \) is the generated disease list.

This pipeline enables the LLM to align its reasoning with high-quality reasoning examples. 
It also facilitates few-shot generalization and improves robustness in complex or ambiguous cases.

\textbf{Related Reasoning Trace.}
To further enhance diagnostic reasoning, we propose a fine-grained retrieval mechanism that operates at the level of each reasoning step. 
We decompose diagnostic reasoning examples into smaller segments and retrieve the most relevant ones based on medical entity overlap.

Concretely, we first segment each \( R_i \) into numbered reasoning steps \( R_{i,j} \). 
Each segment \( R_{i,j} \) is treated as an independent reasoning unit.
We then extract biomedical entities from each \( R_{i,j} \) using a domain-specific named entity recognizer, such as SciSpaCy~\cite{neumann2019scispacy}. Let \( E_{i,j} \) denote the set of entities extracted from reasoning sentence \( R_{i,j} \), and let \( E_C \) denote the set of entities extracted from the input case \( C \). We compute the Jaccard similarity between the two sets:
\begin{equation}
\text{Sim}(C, R_{i,j}) = \frac{|E_C \cap E_{i,j}|}{|E_C \cup E_{i,j}|}
\label{eq:jaccard}
\end{equation}

With this similarity score, we retrieve the top-\( k \) reasoning fragments \( R_{i,j} \) from the database that share the most biomedical entities with the input:
\begin{equation}
\text{TopK}_{\text{entity}}(C) = \text{arg top-}k_{R_{i,j}} \ \text{Sim}(C, R_{i,j})
\label{eq:topk_entity}
\end{equation}

The retrieved reasoning segments are then concatenated and added to the prompt. The final diagnosis is generated by the LLM as:
\begin{equation}
H_{trace} = \text{LLM}(I_{trace}, \text{TopK}_{\text{entity}}(C), C)
\label{eq:final_diag_entity}
\end{equation}

Here, \( I_{trace} \) is a reasoning instruction prompt, and \( H_{trace} \) is the corresponding prediction. This entity-based retrieval approach allows for a fine-grained alignment between the input case and existing reasoning examples, supporting more focused and interpretable diagnostic generation.

\subsubsection{Web Search}
Medical knowledge is continually evolving,  and access to up-to-date information can significantly enhance diagnostic reasoning, particularly in the case of rare diseases. Existing methods often rely on the internal knowledge of LLMs or static knowledge bases, where outdated knowledge may compromise diagnostic accuracy~\cite{wang2025survey}.
To enhance the model’s ability to incorporate real-world and up-to-date medical knowledge, we introduce a web search module.~Following OpenAI's deep research~\cite{openai2025deepresearch,smolagents}, this module enables the model to iteratively retrieve relevant information from the internet and transform them into structured outputs for diagnostic reasoning. We take potential data leakage into consideration by blocking access to possible sources such as PubMed\footnote{\url{https://pubmed.ncbi.nlm.nih.gov/}} and Hugging Face\footnote{\url{https://huggingface.co/}}.

Given an instruction $I_P$ and a clinical case report $C$, the model generates a high-level search plan $P$:

\begin{equation}
P = \text{LLM}(I_P, C)
\end{equation}
where $P$ consists of three components $Q$, $T$, and $N$, representing the search queries, the types of tools to be used (e.g., search, navigation, content extraction), and the estimated number of retrieval steps, respectively:
\begin{equation}
P = (Q, T, N)
\label{eq:plan_components}
\end{equation}
where $Q = \{q_1, \ldots, q_N\}$ and $T = \{t_1, \ldots, t_N\}$.

Then the agent performs a sequence of tool invocations to execute browsing behavior.
\begin{equation}
s_i = \text{Invoke}(t_i, q_i), \quad \text{for } i = 1, \ldots, N
\label{eq:stepwise_tool_invocation}
\end{equation}

At each step $i$, the agent selects tool type $T_i$ (e.g., search engine, navigator, extractor) and applies it using the corresponding keyword query $q_i$. This results in a sequence of interactions $s_1, s_2, \ldots, s_N$, where each $s_i$ represents the retrieved information or browsing action at that step.

Following the complete tool execution trace $S = \{s_1, s_2, \ldots, s_N\}$, the agent interacts with web content by performing actions such as visiting webpages, scrolling through textual material, locating relevant sections, and extracting information.

At each step $i$, the result $s_i$  is used to incrementally update the agent's internal memory state $m_i$:
\begin{equation}
m_i = \text{LLM}(m_{i-1}, s_i)
\label{eq:memory_update}
\end{equation}

The memory $m_i$ accumulates evidence across steps, enabling the agent to refine its understanding and retrieval strategy iteratively. 
Once sufficient evidence has been accumulated in the memory $m_N$, the agent generates a disease list $H_{web}$ based on the retrieved information with a prompt $I_{web}$:
\begin{equation}
H_{web} = LLM(I_{web}, m_N)
\label{eq:memory_update}
\end{equation}

This design allows the agent to operate in open-ended environments, retrieve up-to-date medical knowledge, and support diagnostic predictions based on real-time web content. It is particularly useful in cases where offline knowledge bases are incomplete or outdated.
\subsection{Evidence Integration and Differential Diagnosis}\label{stage2}
Efficient integration of diverse evidence is essential~\cite{llm4rank2025,unified2025}.~As shown in Appendix~\ref{strategy}, simple voting fails to capture and reason over multiple knowledge sources.
In the second stage, we refine the diagnosis by integrating and comparing evidence collected from multiple sources in Stage 1. 
It enhances the robustness and reliability of the final diagnosis. The model outputs a final diagnosis along with a structured reasoning path that reflects the decision process in a clinical manner.

From the previous stage, we obtained four disease lists \( \{H_{web}, H_{SOAP}, H_{case}, H_{trace}\} \), each generated by a distinct medical reasoning module. Each list consists of suspected diseases ordered by likelihood and along with support evidence.

Then we prompt a LLM to perform evidence integration and differential
diagnosis. The process involves the following steps:

1. Disease Matching Across Sources:
   The LLM identifies equivalent terms across the four disease lists (e.g., “myocardial infarction” and “heart attack”), unifying them under canonical names.

2. Support Aggregation via Voting:
   For each unique disease candidate, the model is instructed to count its presence across the four lists and consider its rank if available. This allows the LLM to estimate the degree of inter-model agreement and relative support for each disease.

3. Differential Diagnosis:
   The LLM is then guided to perform analysis between highly ranked candidates (e.g., pneumonia vs. pulmonary embolism), based on the input case’s clinical features and support evidence. The model is encouraged to use clinical logic to promote or demote diseases based on their fit to the case.

4. Final Re-ranking and Output:
   The LLM outputs a disease list and the output is accompanied by a brief explanation or justification per item.

Formally, the process is modeled as:
\begin{equation}
\small
(R,D) = \text{LLM}(I_{multi}, C, H_{web}, H_{SOAP}, H_{case}, H_{trace}),
\end{equation}
where \( I_{multi} \) is the instruction prompt that explicitly asks the model to generate the final prediction.

This integration ensures that comprehensive domain-specific knowledge is considered, improving both diagnostic accuracy and explainability.

%% file: 4Experiments.tex
\section{Experiments}
\begin{table*}[!t]
\centering
\setlength\tabcolsep{6pt}
\begin{tabularx}{\textwidth}{llXXX}
\toprule
\multicolumn{5}{c}{\textbf{MedCaseReasoning Test Set}} \\
\midrule
\textit{Base Models} & \textit{Reasoning Recall} & \textit{H@1 Acc.} & \textit{H@5 Acc.} & \textit{H@10 Acc.} \\
\midrule
DeepSeek-R1              & 0.648
 &0.360	&0.419	&0.442
      \\
Qwen3-14B	&	0.638&0.284	&0.295&	0.295\\
LLaMA-3.1-8B-Instruct$^{\dag}$   & 0.451       & 0.161       & 0.281      & 0.332       \\
Qwen-2.5-7B $^{\dag}$            & 0.324      & 0.174       & 0.252       & 0.287      \\
\midrule
\textit{Fine-Tuned Models} &  &  &  &  \\
\midrule
LLaMA-3.1-8B-Instruct (SFT)$^{\dag}$ & 0.485       & 0.278       & 0.411       & 0.479       \\
Qwen-2.5-7B (SFT) $^{\dag}$       & 0.486       & 0.249       & 0.363       & 0.425     \\
\midrule
\textit{Agentic Methods} &  &  &  &  \\
\midrule
Self-refinement~\cite{madaan2023self} &0.535      &0.336       & 0.462       & 0.496       \\
MedAgents~\cite{tang2024medagents} &0.641      &0.344      & 0.458       & 0.471       \\
OpenAI-DR~\cite{openai2025deepresearch} &0.557      &0.416	&0.553	&0.602
       \\
\name{}\textbf{{\large *}} &\textbf{0.662 }     &\textbf{0.420}      & \textbf{0.577}       & \textbf{0.617}       \\\hline
\multicolumn{5}{c}{\textbf{DiReCT Test Set}}\\
\midrule
 & \textit{Reasoning Recall} & \textit{H@1 Acc.} & \textit{H@5 Acc.} & \textit{H@10 Acc.}\\ 
\hline
DeepSeek-R1 & 0.473&0.293	&0.413	&0.473             \\
Self-refinement~\cite{madaan2023self} &	0.662&0.300&	0.466&	0.586       \\

OpenAI-DR~\cite{openai2025deepresearch} &	0.586&0.297	&0.452&	0.479
       \\
\name{}\textbf{{\large *}} &\textbf{0.665} &\textbf{0.333 }     &\textbf{0.503}      & \textbf{0.587}              \\
\bottomrule
\end{tabularx}
\caption{Performance of models on  test set. Each model is evaluated on reasoning coverage and diagnostic accuracy (broken down into 1, 5, and 10-shot). Results with $^{\dag}$ are reported in the MedCaseReasoning paper~\cite{wu2025medcasereasoning}. Results with \textbf{{\large *}} are averaged over three random runs ($p <
0.05$ under t-test). To ensure a consistent comparison, all agentic methods are implemented based on the DeepSeek-R1 backbone. }
\label{table:medcasereasoning_test_results}
\vspace{-0.3cm}
\end{table*}
\subsection{Experimental Setup}
\subsubsection{Datasets}
We utilize the MedCaseReasoning~\cite{wu2025medcasereasoning} dataset to evaluate diagnostic reasoning methods, as it includes human-annotated reasoning paths that provide gold-standard chains of clinical inference.
The dataset comprises 14,489 diagnostic QA cases, each accompanied by a detailed reasoning statement curated from open-access medical case reports, reflecting real-world clinical thought processes.
We follow the standard split, using 13,092 cases from the training set to construct the case database. For evaluation, due to limited computational resources, we randomly select 300 samples from the test set.

We also conducted experiments on the DiReCT dataset~\cite{wang2024direct}. The dataset consists of clinical notes, each carefully annotated by physicians to reflect the full diagnostic reasoning process from observations to final diagnosis. We randomly selected 50 samples in this dataset for evaluation. 

\subsubsection{Baseline}
To better evaluate the performance of our approach on medical diagnosis tasks, we compare it with three types of baselines:~base models, which represent state-of-the-art 
LLMs of various sizes; fine-tuned models, which are obtained by applying fine-tuning techniques to several base models; and agentic methods.~Specifically, we select several representative agentic methods. Self-Refinement~\cite{madaan2023self} improves reasoning performance through iterative self-optimization in general domain, and has been shown to be effective in medical reasoning~\cite{tang2025medagentsbench} as well. MedAgents is a multi-agent framework specifically designed for medical diagnostic tasks~\cite{tang2024medagents}. OpenAI-DR~\cite{openai2025deepresearch} is a recent agentic method that achieves state-of-the-art performance on complex reasoning tasks. For a fair evaluation, we implement all these agentic methods using DeepSeek-R1 as the backbone.
\subsubsection{Implementation Details}
We use DeepSeek-R1 as the backbone model and utilize the official API to conduct all experiments\footnote{\url{https://www.deepseek.com/}}. Results are averaged over three random runs. While the execution time of our method is subject to fluctuations according to API server or web environment, it typically stays within a minute-level range on average. Prompts used in our experiments are provided in the appendix~\ref{prompt}. We retrieve the top 10 similar cases and reasoning paths in the hierarchical retrieval module. Medical entities are extracted using SciSpacy (version 0.5.5). 
To prevent data leakage, we restrict access to specific websites such as PubMed and Hugging Face within the web search module. Since the DiReCT cases are already formatted in a standardized clinical style, we omit the SOAP structuring module in our pipeline during evaluation on this dataset.

\subsubsection{Evaluation Metrics}
Following MedCaseReasoning~\cite{wu2025medcasereasoning}, we evaluate model performance with reasoning recall and diagnostic accuracy.
Reasoning recall measures the extent to which the model can recover key steps of the clinical reasoning process. 
Hit@k Accuracy (H@1, H@5, H@10) evaluates diagnostic accuracy by measuring whether the correct diagnosis appears in the top-k predictions. 
\subsection{Main Results}

We analyze the performance of our method in comparison with base models, fine-tuned models, and recent agentic approaches.
The results in Table~\ref{table:medcasereasoning_test_results} demonstrate several key observations:

\begin{itemize}[leftmargin=*,nolistsep]
\item Our proposed approach \name{} achieves the best performance across all evaluation metrics. This highlights the effectiveness of our two-stage framework in diagnostic reasoning task.


\item \name{} achieves significant improvements on Hit@5 and Hit@10 metrics.~Specifically, on the MedCaseReasoning test set, it outperforms the backbone model DeepSeek-R1 with improvements of 8.4\%/17.5\%  and surpasses OpenAI-DR by 2.4\%/1.5\% on H@5/H@10. These gains suggest that integrating multi-source knowledge significantly enhances the model’s ability to recall target diagnosis within a broader candidate set.

\item \name{} achieves superior reasoning recall compared to other methods, indicating a better alignment with the ground truth.~This reflects the model’s enhanced capability to reproduce standard diagnostic reasoning processes.
\end{itemize}

\subsection{Ablation Study}
\begin{table}[t]
\centering
\resizebox{0.5\textwidth}{!}{
\begin{tabular}{lcccc}
\toprule
               &$H@1$&  $H@5$&  $H@10$&Recall   \\ \hline
DeepSeek-R1              
 &0.360	&0.419	&0.442&0.648
      \\
w/ SOAP & 0.379	&0.467	&0.502&0.638
 \\
w/ web search     &0.416	&0.553	&0.602&0.460
 \\ 
w/  related case  & 0.393&	0.489&	0.523&0.634
  \\
w/  related trace  &0.386	&0.520	&0.576&0.573
 \\ \hline
\name{}   & \textbf{0.420}	&\textbf{0.577}	&\textbf{0.617}&\textbf{0.662}
  \\ \bottomrule
\end{tabular}}
\caption{Ablation study.}
\label{tab:ablation study}
\vspace{-0.3cm}
\end{table}

To better understand the contribution of each component of \name{}, we experiment with several variants based on DeepSeek-R1.~The results in Table~\ref{tab:ablation study} show the performance of each variant enhanced by a specific type of external knowledge:
\begin{itemize}[leftmargin=*,nolistsep]
\item \textbf{w/ SOAP}: This variant incorporates only structured original cases (SOAP format) without any external knowledge augmentation, yet it still improves upon the base model. It demonstrates that structured clinical data can positively contribute to diagnostic reasoning.

\item \textbf{w/ web search}: 
This variant integrates knowledge obtained through real-time web search and achieves the best performance among all variants.
This result indicates that adaptively incorporating up-to-date external knowledge contributes significantly to accurate diagnosis.


\item \textbf{w/ related case \& w/ related reasoning trace}: 
These two variants retrieve different granularities of knowledge from the case database.
The results show that relevant cases effectively support diagnostic reasoning, especially with improvements in H@5 and H@10, indicating improved recall of the target diagnosis.
The fine-grained retrieval even performs better since it provides more relevant and matching knowledge.
\end{itemize}
Compared to aformentioned variants, \name{} integrates evidence from multiple knowledge sources and achieves the best performance across all metrics.
This demonstrates the effectiveness of comprehensive knowledge integration.

\subsection{Compatibility Study}
To evaluate the generalizability of \name{} across different backbone models, we applied the multi-perspective evidence integration and differential diagnosis steps using Qwen3-14B, and compared the results to agentic methods based on the same backbone. As shown in Table~\ref{table:qwen}, MultiDx consistently improves diagnostic performance across all metrics. The results show that our method remains effective when using smaller-scale LLMs, outperforming other agentic approaches.

\begin{table}[t!]
\centering
\resizebox{1\linewidth}{!}{
\begin{tabular}{lcccc}
\toprule
& $H@1$&  $H@5$&  $H@10$&Recall\\
\hline
Qwen3-14B	&0.284	&0.295&	0.295&	0.638\\
MedAgents&	0.362	&0.377&	0.384&	0.572
\\
Self-refinement &	0.241&	0.345&	0.404&	0.491
\\
MultiDx&	\textbf{0.399}	&\textbf{0.556}	&\textbf{0.601}&	\textbf{0.679}
\\
\bottomrule
\end{tabular}}
\caption{Comparison of Qwen3-14B-based methods on diagnostic reasoning.}
\label{table:qwen}
\vspace{-0.3cm}
\end{table}


\subsection{Seen/Unseen Case Analysis}
\begin{table}[t]
\centering
\setlength\tabcolsep{3pt}
\renewcommand{\arraystretch}{1}
\begin{tabular}{l|cc|cc}
\toprule
\multirow{2}{*}{\textbf{Method}} 

& \multicolumn{2}{c|}{\textbf{Seen}} & \multicolumn{2}{c}{\textbf{Unseen}} \\
& $H@1$&   $H@5$ & $H@1$&    $H@5$ \\
\midrule
DeepSeek-R1              
 &0.459		&0.520	&0.300		&0.366

      \\
w/ SOAP & 0.468&0.572&	0.292&0.367

 \\
w/ web search     &0.511		&0.703&	0.326		&0.413

 \\ 
w/  related case  & 0.489		&0.621&	0.276	&0.365

  \\
w/  related trace  &0.469		&0.647&	0.290	&0.393

 \\ \hline
\name{}   & 0.504	&\textbf{0.710}	&\textbf{0.338}	&	\textbf{0.448}

 \\
\bottomrule
\end{tabular}
\caption{Performance of variants of \name{} on seen and unseen diseases.}
\label{tab:ind}
\vspace{-0.3cm}
\end{table}
To evaluate the generalization ability of different variants, we divide the test set into two subsets: seen cases, where the target diagnosis appears in the case database, and unseen cases, where the disease has not been encountered. Table~\ref{tab:ind} reports the performance of all variants under both settings.

As shown in Table~\ref{tab:ind}, the two variants that retrieve knowledge from the case database perform well on seen cases, indicating the effectiveness of leveraging relevant cases for in-distribution diseases.~The variant that searches knowledge from the web demonstrates improved performance on both seen and unseen samples, highlighting that leveraging comprehensive external knowledge leads to improved generalization and robustness.
Overall, \name{} achieves the best performance across both types of cases, suggesting that it generalizes well to unseen or rare diseases.


\subsection{Computational Cost Analysis}
To better quantify the computational cost of MultiDx, we compared the latency and token usage of each module with two agent-based baselines. Table~\ref{tab:cost} shows the per-case average latency and token consumption across modules.
\begin{table}[t]
\centering
\setlength\tabcolsep{2.pt}
\resizebox{0.5\textwidth}{!}{
\begin{tabular}{lcc}
\toprule
            &  Avg Latency (min) &Avg. Total Tokens    \\ \hline
Self-refinement&	5.3&18,398.2\\	
 OpenAI-DR&	8.0&	N/A	\\
 Stage1-SOAP	&1.6&	4,360.8	\\
 Stage1-Web Search	&8.0&	N/A \\
 Stage1-Related Trace	&6.2&	3,108.0	\\
 Stage1-Related Case	&1.8&	8,591.0	\\
 Stage 2 &	0.46&	4603.7	\\\hline
 MultiDx &	8.46&	\textasciitilde 20,000	
  \\ \bottomrule
\end{tabular}}
\caption{Computational Cost Analysis.}
\label{tab:cost}
\vspace{-0.3cm}
\end{table}
The token usage for the web search module is not reported due to high variability across runs, which depends on real-time network conditions and the number of iterative search steps.

Our all Stage 1 modules can be executed in parallel, which significantly reduces end-to-end latency. Moreover, our method is modular and configurable: for simpler or time-constrained use cases, users can disable high-latency modules such as web search and related trace, reducing the total time to around 2 minutes while still preserving reasonable diagnostic performance.

Compared to agent-based methods such as Self-refinement and OpenAI-DR, our framework demonstrates comparable latency and token consumption, without introducing significant additional cost. Moreover, MultiDx is modular, enabling more controllable and interpretable reasoning, and allowing users to make flexible trade-offs between latency and performance depending on the complexity of the case. 
\subsection{Case Study}
We analyze a representative case to illustrate the model's reasoning process. Specifically, we present the disease lists obtained from different knowledge sources, along with the final answer produced by \name{}. As shown in Table~\ref{tab:case-study0}, the correct answer appears at different ranks across these lists, and \name{} successfully ranks the correct diagnosis at the top after integration.
In addition, our approach provides high-quality reasoning paths. The complete reasoning process is detailed in Appendix~\ref{case}.

\begin{table}[t]
\centering
\small
\begin{tabular}{p{2cm} | p{5cm}}
\toprule
\textbf{SOAP} &
1.Intracranial Germinoma 

2.Metastatic Carcinoma (e.g., from thyroid, lung, or breast primary) 

3.\textbf{Primary Central Nervous System Lymphoma }

...  \\\hline

\textbf{Web Search} &
1.Carcinomatous meningitis

2.Metastatic brain disease (e.g.,
from thyroid, lung, or breast primary) 

3.\textbf{Primary CNS lymphoma} 

... \\\hline
\textbf{RAG-case} &
1.Metastatic disease 

2.\textbf{Primary central nervous system lymphoma} 

... \\\hline
\textbf{RAG-trace} &
1.\textbf{Primary central nervous system lymphoma} 

2.Metastatic disease (e.g., from thyroid carcinoma) 

...   \\
\hline

\textbf{MultiDx}
& 1.\textbf{Primary central nervous system lymphoma} 

2. Neurosarcoidosis 

...
\\\hline
\textbf{Ground Truth}&Primary central nervous system lymphoma\\
\bottomrule
\end{tabular}
\caption{Disease lists from different knowledge sources and the final prediction of \name{} for one case.}
\label{tab:case-study0}
\vspace{-0.3cm}
\end{table}

%% file: 2RelatedWork.tex
\section{Related Work}
\subsection{Agentic Methods}
In recent years, a series of agent-based approaches have significantly enhanced the reasoning capabilities of LLMs~\cite{stepwise2025,align2025,harness2025}. Self-refine~\cite{madaan2023self} improves reasoning quality by prompting the model to revise its own outputs iteratively. Aflow~\cite{zhang2024aflow} optimizes reasoning trajectories by sampling and selecting the most effective workflow structures. Diverging from conventional RAG and agent-based methods, deep research agents have demonstrated strong capabilities on complex and multi-hop tasks by integrating dynamic planning, diverse retrieval tools, and advanced reasoning mechanisms~\cite{huang2025deep}.
Search-o1~\cite{li2025search} proposes a RAG-based agent workflow for dynamic knowledge retrieval, and employs a Reason-in-Documents module that filters and refines retrieved content before injection into the reasoning chain.~OpenAI-DR~\cite{openai2025deepresearch} introduces a adaptive workflow with enhanced multi-modal context processing and integrated toolchains for executing complex tasks.
While above agentic methods have shown strong performance in general reasoning~\cite{bridging2025,navigate2025}, medical reasoning involves unique challenges, such as obtain domain-specific knowledge and follow clinical standards. 
\subsection{Medical Reasoning}
LLMs are widely applied across various downstream tasks~\cite{plate2023,llmemb2025,large2025,rethinking2025,mesh}.~Recent research has explored LLM-based frameworks for medical reasoning. MedAgents~\cite{tang2024medagents} proposes an agentic framework where LLMs take on specialized roles (e.g., cardiology, radiology) and participate in multi-stage discussions to simulate clinical reasoning in real-world. MDAgents~\cite{kim2024mdagents} introduces an adaptive strategy that dynamically allocates agents. ConfAgents~\cite{zhao2025confagents} leverages conformal prediction to filter cases and selectively activate collaborative agents, reducing cost while maintaining diagnostic accuracy. MMedAgent-RL~\cite{xia2025mmedagent} and MedAgent-Pro~\cite{wang2025medagent} extend medical multi-agent frameworks to multimodal settings. MEDDxAgent~\cite{rose2025meddxagent} proposes a modular differential diagnosis system that iteratively refines predictions through simulated history-taking and tool-assisted reasoning. 

However, most of aforementioned frameworks fall short in integrating diverse knowledge sources, particularly in dynamically retrieving information from real-time web data, which limits their capability for effective medical reasoning. In addition, they often focus on final results while neglecting the assessment of the reasoning process.

%% file: 5Conclusion.tex
\section{Conclusion}
In this paper, we propose \name{}, a two-stage diagnostic reasoning framework enhanced by multi-source knowledge. It integrates evidence from diverse knowledge sources to generate accurate diagnostic predictions as well as high-quality reasoning trajectories.~Extensive experiments demonstrate the effectiveness of the proposed method.

%% file: 6Limitations.tex
\section*{Limitations}
While \name{} effectively integrates multiple sources of knowledge, its performance may be affected if the external sources are of low quality or contain noise. Moreover, the current two-stage design decouples knowledge extraction and differential diagnosis. In the future, we plan to explore joint learning mechanisms to improve overall coherence and performance.

\section*{Acknowledgments}
This work is in part funded by the National Key Research and Development
Program of China (2023YFC3321600); in part by National Natural Science Foundation of China (Grant No. 62372364) and the Technical Innovation Guidance Plan of Shaanxi Province, China (Grant No. 2024QCY-KXJ-199); in part by  National Natural Science Foundation of China (No.62502404), Hong Kong Research Grants Council (Research Impact Fund No.R1015-23, Collaborative Research Fund No.C1043-24GF, General Research Fund No. 11218325), Institute of Digital Medicine of City University of Hong Kong (No.9229503), and Tencent (Tencent Rhino-Bird Focused Research Program, Tencent University Cooperation Project).

%% file: 7Appendix.tex
\section{In-depth Analysis}
\subsection{Integration Strategy}\label{strategy}
\begin{table}[t]
\centering
\setlength\tabcolsep{9pt}
\begin{tabular}{lccc}
\toprule
               & $H@1$&  $H@5$&  $H@10$    \\ \hline

w/ vote  &0.403	&0.552	&0.604

 \\ 
w/ D.D.   & \textbf{0.42}	&\textbf{0.577}	&\textbf{0.617}
  \\ \bottomrule
\end{tabular}
\caption{Performance of different integration strategy used in stage2.}
\label{tab:stage2}
\end{table}
To investigate the impact of different integration strategies, we compare simple voting~(w/ vote) with our proposed stage 2: Evidence Integration and Differential Diagnosis (w/ D.D.).

As shown in the Table~\ref{tab:stage2}, although the simple voting provides a straightforward way to select the most frequent answer, it lacks the ability to analysis the medical evidence of each candidate within the given context.
In contrast, our approach explicitly prompts the model to compare different diagnoses, enabling more informed and reasonable decisions.


\subsection{Parameter Analysis}
\begin{table}[t!]
\centering
\resizebox{1\linewidth}{!}{
\begin{tabular}{lccc}
\toprule
\textbf{Retrieved Sample}& $H@1$&  $H@5$&  $H@10$\\
\hline
\textbf{$k=0$} &0.360 &0.419 &0.442\\
\textbf{$k=2$}& 0.368&	\textbf{0.502}&	\textbf{0.559}
\\
\textbf{$k=5$}& 0.380&	0.491&	0.540
\\
\textbf{$k=10$}& \textbf{0.393}&	0.489	&0.523
\\
\textbf{$k=15$}&0.391	&0.490	&0.529
\\
\bottomrule
\end{tabular}}
\caption{Effect of the number of retrieved medical cases ($k$) on diagnosis accuracy in the case-enhanced Generation Module. Performance is measured by Hit@1, Hit@5, and Hit@10.}
\label{table:multiqa_retrievers}
\vspace{-0.3cm}
\end{table}

To evaluate the parameter sensitivity of \name{}, we investigate the impact of the number of retrieved medical cases ($k$) on diagnostic accuracy within the case retrieval module. 
As shown in Table~\ref{table:multiqa_retrievers},
increasing $k$ from 2 to 10 leads to an improvement in H@1, indicating that retrieving more cases helps the model make more accurate diagnosis predictions.
However, H@5 and H@10 slightly decrease during the same range, suggesting that the inclusion of additional cases may introduce noise or irrelevant information, which hinders the recall of correct diagnosis.
As $k$ increases from 10 to 15, the performance remains relatively stable.

Overall, incorporating retrieved cases improves diagnostic performance, and the model demonstrates robustness to $k$ within a reasonable range.


\section{Prompt}\label{prompt}
In this section, we provide additional experimental details to facilitate better reproducibility. In Table~\ref{tab:prompt1}, \ref{tab:prompt2}, we present example prompts used in each module of \name{}. Since web search module involves a larger number of prompts, detailed prompt examples of this module are provided in our code repository due to space limitations~\footnote{\url{https://github.com/Applied-Machine-Learning-Lab/ACL2026-MultiDx}}.
Our output and evaluation templates follow those of MedCaseReasoning~\cite{wu2025medcasereasoning} to ensure consistency.

\section{Case Study}\label{case}
In this section, we conduct a case study, using a specific example to explicitly illustrate the output of each module, including the list of suspected diseases and the final reasoning trajectory along with the diagnosis prediction. This example demonstrates how our method performs diagnostic reasoning in a step-by-step manner.

Table~\ref{tab:case-study} illustrates how individual evidence sources contribute to differential diagnosis generation. 
Table~\ref{tab:case-study2} presents a side-by-side comparison between the ground truth reasoning trajectory (from the MedCaseReasoning dataset, validated by expert annotators) and the output reasoning results from MultiDx for a complex CNS case. The two show strong consistency in terms of core diagnostic hypotheses, reasoning structure, exclusion logic, and evidence grounding.

Based on this case, we conducted a qualitative analysis to further validate how MultiDx's reasoning trace aligns with expert clinical thinking. Specifically, MultiDx correctly identifies primary CNS lymphoma as key differential diagnose, supported by clinical findings such as CSF pleocytosis, elevated IgG, and multifocal brain lesions. The model organizes its evidence by diagnostic categories (e.g., neoplastic, infectious, inflammatory) and provides clear justification for prioritization. It also demonstrates well-considered clinical judgment by excluding less likely options (e.g., metastatic disease due to lack of primary tumor), reflecting a degree of uncertainty handling similar to human experts.

To further enhance transparency, we have collected and organized the complete reasoning traces from all four modules, as shown in Table~\ref{tab:case-study3},\ref{tab:case-study4}.

\begin{table*}[t]
\centering
\renewcommand{\arraystretch}{1.3}
\setlength{\tabcolsep}{10pt}
\begin{tabular}{p{4cm} | p{10cm}}
\hline

\textbf{$I_\text{SOAP}$} &
  Read the following case presentation and give the most likely diagnosis.
First, please classify all extracted facts into the following categories:

---

**S — Subjective**  
Patient-reported symptoms, complaints, history of present illness, medication use, and relevant negatives (e.g., "denies fever"). Include time-related info if mentioned.

**O — Objective**  
Clinician-observed findings: physical exam results, lab/imaging findings, vital signs, and other measurable or observable data.

**A — Assessment**  
If any differential diagnoses, impressions, or suspected conditions are mentioned in the text, extract them here.

**P — Plan**  
If the text includes management plans, tests ordered, treatments started, or follow-up instructions, include them here.

---

Then, provide your internal reasoning for the diagnosis within the tags <think> ... </think>.

Finally, output the final diagnosis (just the name of the disease/entity) within the tags <answer> ... </answer>.
What are the top 10 most likely diagnoses? Be precise, listing one diagnosis per line, and try to cover many unique possibilities (at least 10).
----------------------------------------

CASE PRESENTATION

----------------------------------------

<case>

  \\\hline
$I_{case}$ &
Read the following case presentation and give the most likely diagnosis.

What are the top 10 most likely diagnoses? Be precise, listing one diagnosis per line, and try to cover many unique possibilities (at least 10).
Here are some reasoning examples:
<top-k cases>

----------------------------------------

CASE PRESENTATION

----------------------------------------

<case>
 \\\hline

\end{tabular}
\caption{Prompt.}
\label{tab:prompt1}
\end{table*}

\begin{table*}[t]
\centering
\renewcommand{\arraystretch}{1.3}
\setlength{\tabcolsep}{10pt}
\begin{tabular}{p{4cm} | p{10cm}}
\hline

$I_{trace}$ &
   
Read the following case presentation and give the most likely diagnosis.

What are the top 10 most likely diagnoses? Be precise, listing one diagnosis per line, and try to cover many unique possibilities (at least 10).

Some possible disease-related symptoms are listed (optional to use).

<tok-k reasoning paths>

----------------------------------------

CASE PRESENTATION

----------------------------------------

<case>
\\
$I_{multi}$&You are a medical reasoning expert. You are given a clinical case prompt along with diagnostic reasoning traces and suspected disease lists from four different medical reasoning systems.

Your task is to:

1. Carefully consider the case prompt.

2. Analyze and consolidate the reasoning traces from the four sources.

3. Based on the reasoning and the case details, produce:

   - A coherent and medically sound final reasoning trace.
   
   - A ranked list of suspected diseases, ordered by confidence (from highest to lowest), based on the degree of support across sources.
   
[Deep Search Reasoning Trace]

{pred\_search}

[SOAP Reasoning Trace]

{pred\_SOAP}

[RAG Reasoning Trace]

{pred\_trace}

[RAG Reasoning Trace]

{pred\_rag}

----------------------------------------

CASE PRESENTATION

----------------------------------------

<case>
\\
\hline
\end{tabular}
\caption{Prompt.}
\label{tab:prompt2}
\end{table*}

\begin{table*}[t]
\centering
\renewcommand{\arraystretch}{1.3}
\setlength{\tabcolsep}{10pt}
\begin{tabular}{p{4cm} | p{10cm}}
\hline
\textbf{Case} &
\textit{A 70-year-old woman with atrial fibrillation on enoxaparin, liver cirrhosis, type 2 diabetes mellitus, hypertension, and a history of venous thrombus presented with a 5-day history of left arm weakness, numbness in the left fourth and fifth digits and medial palmar surface, and confusion. She reported chronic vision loss in the right eye but denied headache or trauma. On examination, she was alert; oriented to person, place, and time; and disoriented to situation. Visual acuity was reduced in the right eye. Muscle strength was 4/5 in the left triceps, wrist flexors, wrist extensors, and finger flexors, and 3/5 in the left finger extensors. The left triceps reflex was absent. Sensation to pinprick and light touch was decreased in the left fourth and fifth digits. There was no dysmetria or dysdiadokokinesia. ...
} 
\\
\textbf{Answer}&Primary central nervous system lymphoma\\
\hline
\textbf{SOAP} &
1. Intracranial Germinoma 2. Metastatic Carcinoma (e.g., from thyroid, lung, or breast primary) 3. \textbf{Primary Central Nervous System Lymphoma }4. Neurosarcoidosis 5. Leptomeningeal Carcinomatosis 6. Multiple Sclerosis 7. Autoimmune Encephalitis (e.g., anti-NMDA receptor encephalitis) 8. Cerebral Vasculitis (e.g., primary angiitis of the CNS) 9. Infectious Meningoencephalitis (e.g., viral, fungal, or tuberculous) 10. Paraneoplastic Neurological Syndrome  \\

\textbf{Web Search} &
1. Carcinomatous meningitis 2. Metastatic brain disease (e.g., from thyroid, lung, or breast primary) 3. \textbf{Primary CNS lymphoma} 4. Neurosarcoidosis 5. Chronic meningitis of unknown etiology 6. Lymphomatous meningitis 7. Tuberculous meningitis 8. Fungal meningitis 9. Autoimmune encephalitis 10. Vasculitis of the central nervous system  \\
\textbf{RAG-case} &
1. Metastatic disease 2. \textbf{Primary central nervous system lymphoma} 3. Neurosarcoidosis 4. Tuberculous meningitis 5. Fungal meningitis 6. Leptomeningeal carcinomatosis 7. Glioblastoma multiforme 8. Inflammatory pseudotumor 9. IgG4-related disease 10. Atrial myxoma with cerebral metastases  \\
\textbf{RAG-trace} &
1. \textbf{Primary central nervous system lymphoma} 2. Metastatic disease (e.g., from thyroid carcinoma) 3. Germ cell tumor (e.g., germinoma) 4. Neurosarcoidosis 5. Meningiomatosis or multiple meningiomas 6. Tuberculosis or other chronic infectious meningitis 7. Demyelinating disease (e.g., multiple sclerosis) 8. Vascular malformation or aneurysm 9. Pineal parenchymal tumor (e.g., pineocytoma or pineoblastoma) 10. Leptomeningeal carcinomatosis from occult primary   \\
\hline

\textbf{MultiDx}
& 1. \textbf{Primary central nervous system lymphoma} 2. Neurosarcoidosis 3. Metastatic brain disease (e.g., from thyroid or unknown primary) 4. Carcinomatous meningitis 5. Germ cell tumor (e.g., germinoma) 6. IgG4-related disease 7. Tuberculous meningitis 8. Fungal meningitis 9. Leptomeningeal carcinomatosis 10. Autoimmune encephalitis 
\\

\hline
\end{tabular}
\caption{Case study.}
\label{tab:case-study}
\end{table*}

\begin{table*}[t]
\centering
\small
\renewcommand{\arraystretch}{1.3}
\setlength{\tabcolsep}{10pt}
\begin{tabular}{p{3cm} | p{10cm}}
\hline

\textbf{Ground Truth} &
1. Embolic stroke was considered given her atrial fibrillation and diabetes but felt less likely — “While the patient had several risk factors for embolic stroke, including atrial fibrillation and diabetes, the specific patterns of sensory loss in the left upper extremity and loss of the left triceps reflex were more suggestive of either nerve root or brachial plexus pathology.” 2. Extradural or intradural extramedullary spinal canal mass and idiopathic brachial neuritis were considered given her focal arm findings — “Our differential diagnosis included a lateralized mass in the extradural or intradural extramedullary spinal canal or idiopathic brachial neuritis.” 3. Leptomeningeal metastases from carcinoma or lymphoma were considered after CN III involvement — “Considerations included leptomeningeal disease from metastatic carcinoma or lymphoma as well as PCNSL...” 4. Primary central nervous system lymphoma was considered due to her age and multifocal neurologic presentation — “Considerations included...PCNSL due to the patient’s age and multifocal nature of her symptoms...” 5. Tuberculosis, histoplasmosis, coccidioidomycosis, and cryptococcosis were considered among infectious etiologies — “Considerations included...tuberculosis, histoplasmosis, coccidioidomycosis, or cryptococcosis...” 6. Sarcoidosis was considered among inflammatory causes — “Considerations included...sarcoidosis.” 7. Neurolymphomatosis of the brachial plexus was considered but not supported by imaging — “Gadolinium-enhanced brachial plexus and spine MRI were unrevealing.” 8. Metastatic thyroid carcinoma was considered given the CT findings — “Chest/ abdomen/pelvis CT with contrast demonstrated multiple nodules in the thyroid concerning for metastases.” 9. Leptomeningeal dissemination and infection were less likely after nondiagnostic CSF studies — “CSF studies from two serial lumbar punctures...demonstrated lymphocytic pleocytosis, elevated protein, and elevated IgG index but neither malignant cells nor other inflammatory and infectious markers.”
   \\
\hline

\textbf{MultiDx}
& 
After consolidating the reasoning traces from the four sources, several key points emerge: - The multifocal CNS lesions, rapid progression, and CSF abnormalities strongly suggest an aggressive neoplastic or inflammatory process. 

- \textbf{Primary CNS lymphoma} is highly supported due to its typical presentation with multifocal enhancing lesions, CSF pleocytosis, elevated IgG index, and occurrence in older adults. The lack of malignant cells in CSF does not rule it out entirely, as cytology can be negative. 

- \textbf{Neurosarcoidosis} is also well-supported, given the cranial nerve involvement, dural lesions, elevated IgG index, and potential systemic correlates (e.g., thyroid nodules could represent granulomas). However, the rapid progression might be less typical. 

- \textbf{Metastatic disease} is plausible due to the thyroid nodules and brain lesions, but the absence of a confirmed primary and negative CSF cytology reduces its likelihood. Brain metastases from thyroid cancer are rare. 

- Other considerations include \textbf{carcinomatous meningitis, germ cell tumors, and inflammatory disorders}, but they have less consistent support across sources. 

- \textbf{Infectious etiologies} like tuberculous or fungal meningitis are less likely due to negative markers and lack of systemic signs. 

- The elevated IgG index points towards \textbf{inflammatory or autoimmune} processes, but the aggressive course favors neoplasms. Based on the degree of support across sources and medical plausibility, primary CNS lymphoma is the most likely diagnosis, followed by neurosarcoidosis. The ranked list prioritizes conditions that best explain the multifocal lesions, CSF findings, and progression.
\\

\hline
\end{tabular}
\caption{Case study.}
\label{tab:case-study2}
\end{table*}

\begin{table*}[t]
\centering
\small
\renewcommand{\arraystretch}{1.3}
\setlength{\tabcolsep}{10pt}
\begin{tabular}{p{3cm} | p{10cm}}
\hline

\textbf{SOAP} &
The patient presents with multifocal neurological deficits, including left arm weakness and numbness (suggesting brachial plexus or root involvement), confusion, and a right third cranial nerve palsy that progresses. Imaging reveals lesions in characteristic locations such as the suprasellar and pineal regions, which are common sites for germ cell tumors like germinoma. The CSF shows inflammatory changes (lymphocytic pleocytosis, elevated protein, and IgG index) without evidence of infection or malignancy, which can occur with germinomas due to their inflammatory response. However, the patient's age of 70 is atypical for germinoma, which is more common in younger adults. Alternative considerations include metastatic cancer (given the thyroid nodules suspicious for metastases), primary CNS lymphoma, or neurosarcoidosis. The dural thickening and multifocal enhancing lesions favor a neoplastic process, but the inflammatory CSF and absence of malignant cells might support germinoma or an autoimmune disorder. The rapid progression and systemic findings (thyroid nodules) weigh towards a malignant etiology, but the specific lesion locations make germinoma a strong candidate despite the age.
   \\
\hline

\textbf{Web Search}
& 
The case involves a 70-year-old woman with a history of atrial fibrillation on enoxaparin, liver cirrhosis, type 2 diabetes mellitus, hypertension, and venous thrombus. She presented with acute neurological symptoms including left arm weakness, numbness in specific digits, confusion, and later developed a right third cranial nerve palsy. Imaging showed a hyperdense suprasellar lesion on CT, and MRI revealed enhancing lesions in suprasellar, pineal, and periatrial regions with interval enlargement and dural thickening. CT abdomen showed thyroid nodules concerning for metastases. CSF analysis indicated lymphocytic pleocytosis, elevated protein, and elevated IgG index without malignant cells or infectious markers. This clinical picture is highly suggestive of a neoplastic process with CNS involvement, most likely carcinomatous meningitis, given the multifocal neurological signs, CSF findings, and possible primary malignancy. Alternative diagnoses include metastatic disease, lymphoma, or inflammatory disorders, but the rapid progression and lack of infectious markers favor a neoplastic etiology.
\\

\hline
\end{tabular}
\caption{Case study.}
\label{tab:case-study3}
\end{table*}

\begin{table*}[t]
\centering
\small
\renewcommand{\arraystretch}{1.3}
\setlength{\tabcolsep}{10pt}
\begin{tabular}{p{3cm} | p{10cm}}
\hline

\textbf{Related Case} &1. Pituitary adenoma – Considered due to the suprasellar lesion on CT, but the multifocal nature and rapid progression argue against a typical adenoma.

2. Meningioma – The dural thickening could suggest meningioma, but the multifocal and heterogeneously enhancing lesions with rapid growth are atypical.

3. Metastatic disease – Highly likely given the thyroid nodules concerning for metastases and multiple brain lesions; however, no primary was confirmed and CSF lacked malignant cells.

4. Lymphoma – Primary CNS lymphoma can present with multifocal enhancing lesions and CSF pleocytosis, but the thyroid involvement is unusual.

5. Neurosarcoidosis – Could explain multifocal lesions and CSF findings, but the rapid progression and thyroid nodules are not typical.

6. Tuberculosis or fungal infection – Chronic infections can cause basilar enhancement and CSF abnormalities, but there were no systemic signs of infection and markers were negative.

7. Aneurysmal subarachnoid hemorrhage – The initial hyperdense suprasellar lesion could represent a thrombosed aneurysm, but the subsequent multifocal enhancing lesions are not consistent.

8. Inflammatory or autoimmune disorders – Such as CLIPPERS or IgG4-related disease, could account for the lesions and CSF findings, but the thyroid nodules are atypical.

9. Glioblastoma multiforme – Can be multifocal and show rapid growth, but the suprasellar and pineal locations are uncommon.

10. Atrial myxoma with metastases – Given her atrial fibrillation, cardiac myxoma could embolize or metastasize to brain, but no cardiac mass was reported.
   \\
\hline

\textbf{Related Trace}
& 
The patient is a 70-year-old woman with multiple intracranial enhancing lesions in the suprasellar, pineal, and right periatrial regions, along with dural thickening. The CSF findings show lymphocytic pleocytosis, elevated protein, and an elevated IgG index, which are indicative of an inflammatory or neoplastic process. Systemic imaging (CT chest/abdomen/pelvis) did not reveal a primary malignancy elsewhere, except for thyroid nodules that are concerning for metastases, but these could be incidental given their common occurrence in elderly patients. The neurological symptoms, including cranial nerve palsy, weakness, and sensory deficits, along with the progression of lesions on MRI, strongly suggest a neoplastic etiology. Primary CNS lymphoma is a leading consideration due to its propensity for multifocal enhancing lesions, CSF abnormalities, and occurrence in older adults. Other possibilities include metastatic disease, germ cell tumors, or inflammatory conditions like neurosarcoidosis, but the absence of infectious markers and the rapid progression make lymphoma most likely. The thyroid nodules might represent a primary source, but brain metastases from thyroid cancer are rare and typically not in these locations, reducing its likelihood compared to CNS lymphoma.
\\

\hline
\end{tabular}
\caption{Case study.}
\label{tab:case-study4}
\end{table*}